\documentclass[twocolumn]{article}
\usepackage{amsmath, amssymb, graphicx, subcaption, booktabs}
\usepackage{pgfplots}
\pgfplotsset{compat=1.18}

\DeclareMathOperator{\sech}{sech}

\AtBeginEnvironment{table}{%
  \scriptsize                
  \setlength{\tabcolsep}{3pt}%
}

\usepackage{colortbl}
\newcommand{\revised}[1]{\textcolor{black}{#1}}

\usepackage{hyperref}
\usepackage{multirow}
\hypersetup{
    colorlinks=true,
    linkcolor=blue,
    citecolor=blue,
    urlcolor=blue
}

\newcommand{\method}{VeLU}

\title{\revised{\textbf{\method{}}: \textbf{V}ariance-\textbf{e}nhanced \textbf{L}earning \textbf{U}nit} for Deep Neural Networks}

\author{
     Ashkan Shakarami$^{1,2}$\thanks{Corresponding author: ashkan.shakarami@uni-bonn.de, ashkan.shakarami.ai@gmail.com},  
     Yousef Yeganeh$^{3}$,  
     Azade Farshad$^{3}$,  
     Lorenzo Nicolè$^{2}$,\\
     Stefano Ghidoni$^{2}$,  
     Nassir Navab$^{3}$ \\ 
     \small 
     $^1$University of Bonn, Germany\\
     \small
     $^2$University of Padova, Italy \\
     \small
     $^3$Technical University of Munich, Germany \\
      \small\texttt{\{ashkan.shakarami, lorenzo.nicole\}@phd.unipd.it, stefano.ghidoni@unipd.it} \\
     \small\texttt{\{y.yeganeh, azade.farshad, nassir.navab\}@tum.de}
}

\date{}
\begin{document}
\maketitle
\begin{abstract}
\small 
Activation functions play a critical role in deep neural networks by shaping gradient flow, optimization stability, and generalization. While ReLU remains widely used due to its simplicity, it suffers from gradient sparsity and dead-neuron issues and offers no adaptivity to input statistics. Smooth alternatives such as Swish and GELU improve gradient propagation but still apply a fixed transformation regardless of the activation distribution. In this paper, we propose \textbf{\method{}}, a Variance-enhanced Learning Unit that introduces variance-aware and distributionally aligned nonlinearity through a principled combination of ArcTan-ArcSin transformations, adaptive scaling, and Wasserstein-2 regularization (Optimal Transport). This design enables \method{} to modulate its response based on local activation variance, mitigate internal covariate shift at the activation level, and improve training stability without adding learnable parameters or architectural overhead. Extensive experiments across six deep neural networks show that \method{} outperforms ReLU, ReLU6, Swish, and GELU on 12 vision benchmarks. The implementation of \method{} is publicly available in \href{https://github.com/AshkanShakarami/Velu}{GitHub}.

\end{abstract}

\section{Introduction}

Activation functions are key building blocks in deep neural networks, as they determine the stability of training, the effectiveness of gradient propagation, and the model’s ability to generalize. The Rectified Linear Unit (ReLU)~\cite{nair2010rectified, krizhevsky2012imagenet} remains widely used due to its simplicity and computational efficiency. However, ReLU suffers from the dying-neuron problem, where units with persistently negative preactivations stop contributing gradients during training. Several variants such as Leaky ReLU~\cite{maas2013rectifier}, PReLU~\cite{he2015delving}, and ELU~\cite{clevert2015fast} alleviate this limitation by allowing non-zero gradients for negative inputs.

More recent activation functions, including Swish~\cite{ramachandran2017searching} and GELU~\cite{hendrycks2016gaussian}, introduce smooth, non-monotonic behavior and improve gradient flow. However, these functions apply a fixed transformation independent of the statistical properties of the input, limiting their adaptability across layers where activation distributions may vary considerably. Adaptive activation approaches such as Dynamic ReLU~\cite{chen2020dynamic} attempt to address this through input-conditioned or learnable parameters, but at the cost of increased architectural or computational overhead.

Although numerous variants of ReLU, Swish, and adaptive nonlinearities have been proposed~\cite{agarap2018deep, ramachandran2017searching}, a key open question remains: \textit{\textbf{can we design an activation function that provides adaptivity, non-linearity, and distributional stability - without adding trainable parameters or modifying model architecture?}}  
To this end, we introduce \method{}, a Variance-enhanced Learning Unit that adaptively modulates its response based on the local variance of input activations, enabling data-dependent nonlinearity while preserving the plug-and-play nature of standard activation layers.

Unlike conventional activation functions, \method{} directly mitigates internal covariate shift~\cite{ioffe2015batch} by incorporating variance-aware scaling and distributional alignment at the activation level, rather than relying on batch-wide or feature-wide normalization statistics. This property enhances training stability in deep Convolutional Neural Networks (CNNs) and Vision Transformers (ViTs), where activation distributions can drift substantially during optimization.

The design of \method{} is grounded in a principled composition of smooth transformations. The ArcTan-ArcSin component introduces controlled curvature and bounded non-linearity, improving gradient behavior while avoiding saturation. A variance-based modulation term dynamically scales the activation based on local activation statistics, enabling continuous adaptivity without requiring additional learnable parameters. Finally, a Wasserstein-2 regularization component, inspired by Optimal Transport theory~\cite{arjovsky2017wasserstein, courty2017optimal, fatras2023optimal}, softly aligns activation outputs with a target Gaussian distribution, stabilizing representation dynamics and reducing activation drift (While OT-based methods have been widely used in generative modeling and domain adaptation~\cite{gulrajani2017improved}, their integration into activation design remains largely unexplored).

Extensive experiments on CIFAR-10, CIFAR-100, MNIST, Fashion-MNIST, Corel-1K, and Corel-10K, PathMNIST, BloodMNIST, PACS, EuroSAT-RGB, Oxford-IIIT Pet, and Tiny-ImageNet demonstrate that \method{} consistently improves performance across six architectures, including ViT-B16, VGG19, ResNet50, DenseNet121, MobileNetV2, and EfficientNetB3~\cite{sandler2018mobilenetv2, huang2017densely, shakarami2021fast, shakarami2025depvit, shakarami2025transformative}. These results position \method{} as an effective, architecture-agnostic, and computationally lightweight alternative to conventional activation functions for deep neural networks.

\section{\method{}}
\subsection{Structure}
\label{sec:Tunable_Parameters}

Mathematically, \method{} is defined in \autoref{eq:velu}, where $\sigma(x)$ denotes the sigmoid function and $\lambda$ is a trainable scaling parameter. The hyperparameters $\alpha$, $\beta_1$, $\beta_2$, $\gamma$, $\mu$, the momentum term $m$, and the Wasserstein weight $\lambda_{OT}$ jointly govern the curvature, adaptivity, and distributional behavior of the activation. Importantly, \method{} adds no new layers and only one trainable scalar ($\lambda$), preserving the architectural simplicity of standard activation layers while enabling variance-aware modulation. In the following, we provide a detailed description and motivation for each component.

\begin{equation}
\label{eq:velu}
\begin{aligned}
f(x) &= \lambda.\, x. \,
\sigma\!\left(
    \alpha \left[ \tan^{-1}(\beta_1 x) + \sin^{-1}(\beta_2 x) \right]
\right)
\\
&\quad \times \left( 1 + \gamma \tanh(m\,\mu\,\sigma_x) \right)
- \lambda_{OT}.\, W_2\!\left(x, \mathcal{N}(0,1)\right).
\end{aligned}
\end{equation}

\paragraph{\textbf{Motivation for the components.}}
Following the design rationale, each term plays a specific role: 

\begin{itemize}

    \item \textbf{Scaling factor $\lambda$:\\} A learnable scalar that lets the network adjust the global amplitude of the activation to compensate for the composite transformation.
    
    \item \textbf{Base term $x$:\\} Maintains an identity-like behavior near the origin, preserving gradient flow when inputs are small.
    
    \item \textbf{ArcTan-ArcSin composite:\\} \(
       \sigma\!\big(\alpha(\tan^{-1}(\beta_1 x)+\sin^{-1}(\beta_2 x))\big)
    \) provides smooth, bounded non-linearity. The combination of $\tan^{-1}$ and $\sin^{-1}$ yields gradual saturation and symmetric curvature, mitigating abrupt plateaus seen in hard-saturating or piecewise-linear units. Scaling Parameter $\alpha$ controls the strength of the non-linearity by scaling the ArcTan-ArcSin argument before the sigmoid. Larger $\alpha$ increases curvature and sharp transitions; smaller values lead to smoother behavior. Transformation Parameters ($\beta_1$ and $\beta_2$ scale the inputs to arctan and arcsin, respectively. Larger values accentuate non-linear regions and produce sharper responses; smaller values make the activation more linear. This tunability allows fine-grained control over the shape of the gating non-linearity.
    
    \item \textbf{Variance-adaptive modulation:\\} 
    \(
       1 + \gamma \tanh(m \mu \sigma_x)
    \)
    adapts the effective non-linearity to the dispersion of activations. The hyperbolic tangent keeps the factor bounded, while $m$ controls how sensitively this term reacts to changes in $\sigma_x$, similar in spirit to a “momentum” on variance. Adaptive scaling factor $\gamma$ determines how strongly the activation is modulated by variance. Larger $\gamma$ increases the amplitude of variance-driven adaptation; very large values can over-amplify high-variance regions and thus require care. The momentum hyperparameter $m$ adjusts the sensitivity of the variance-aware scaling to changes in $\sigma_x$ and acts as a smoothing factor over the batch statistics.
    
    \item \textbf{Wasserstein-2 regularization:\\} 
    \(
      -\lambda_{OT} W_2(x,\mathcal N(0,1))
    \)
    softly aligns the activation statistics with a Gaussian target, promoting stability and mitigating activation drift in a geometry-aware way.  $\lambda_{OT}$ controls the contribution of the Wasserstein-2 penalty. Larger $\lambda_{OT}$ enforces stronger statistical alignment with the Gaussian prior, which can improve stability and generalization but may hamper convergence if chosen too large.

\end{itemize}

\subsection{Visual behavior}

\begin{figure}[t]
    \centering
    \includegraphics[width=0.48\textwidth]{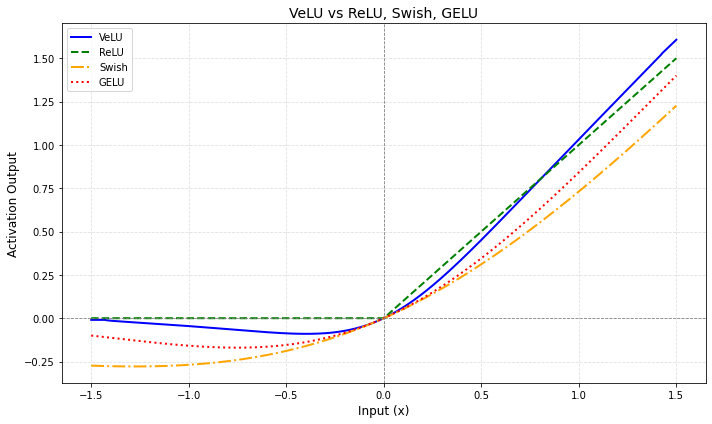}
    \caption{Comparison of \method{} with standard activation functions (ReLU, Swish, GELU) in the range $[-1.5, 1.5]$. The plot highlights smoothness, non-linearity, and the adaptive behavior introduced by \method{}.}
    \label{fig:VeluVsAll}
\end{figure}

\autoref{fig:VeluVsAll} compares \method{} with standard activation functions (ReLU, Swish, GELU) over the input interval $[-1.5, 1.5]$. ReLU exhibits a sharp kink at $x=0$ and zero response for negative inputs, which can lead to dead units. Swish and GELU provide smooth transitions and improved gradient flow, yet they apply a static transformation that does not adapt to activation statistics. In contrast, \method{} maintains smooth, bounded behavior while introducing self-adaptive scaling and distributional regularization via the Wasserstein-2 term. This comparison highlights the foundational curvature and adaptivity advantages of \method{} over traditional activations.

\begin{figure}[t]
    \centering
    \includegraphics[width=0.48\textwidth]{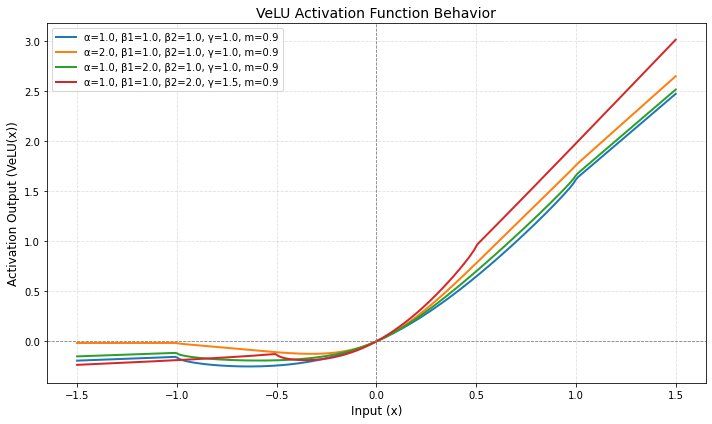}
    \caption{Behavior of \method{} under different parameter configurations. The curves illustrate how curvature, smoothness, and variance-adaptive modulation emerge from the ArcTan-ArcSin composite, scaling term, and Wasserstein alignment.}
    \label{fig:velu_activation_behavior}
\end{figure}

Building on this comparison, \autoref{fig:velu_activation_behavior} visualizes how the internal components of \method{} interact under different configurations of $\alpha$, $\beta_1$, $\beta_2$, $\gamma$, $\mu$, and $m$. The ArcTan-ArcSin composite governs curvature and smoothness, while the variance-based term $(\sigma_x)$ dynamically adapts the response to the dispersion of preactivations. Together, these mechanisms yield a richer and more flexible activation landscape than static nonlinearities. The Wasserstein-2 regularization term further encourages the activation outputs to align with a stable Gaussian target, reducing activation drift and mitigating internal covariate shift at the activation level.

\subsection{Mathematical characteristics}

\method{} exhibits a non-monotonic yet smooth response due to the ArcTan-ArcSin structure. Unlike ReLU, which thresholds negative values at zero, \method{} retains small negative activations, helping to propagate gradients and avoid neuron death.

The variance-based scaling factor is defined as
\begin{equation}
\label{eq:ArcTan-Sin_formulation}
s_{\text{adaptive}} = 1 + \gamma \cdot \tanh(m \cdot \mu \cdot \sigma_x),
\end{equation}
where $\sigma_x$ is the standard deviation of preactivations over the mini-batch. This factor adjusts the effective non-linearity according to activation dispersion: for small $\sigma_x$, $s_{\text{adaptive}}\approx 1$; for large $\sigma_x$, it approaches $1+\gamma$ (with saturation mediated by $m\cdot \mu$).

To further stabilize training, \method{} incorporates Wasserstein-2 distance minimization between its outputs and a Gaussian target:
\begin{equation}
\label{eq:Wasserstein-2_distance_minimization}
W_2^2 = (E[x] - \mu)^2 + (\sigma_x - \sigma_{\text{target}})^2,
\end{equation}
which penalizes deviations in mean and variance, encouraging well-conditioned internal representations.

\begin{figure}[t]
    \centering
    \includegraphics[width=0.5\textwidth]{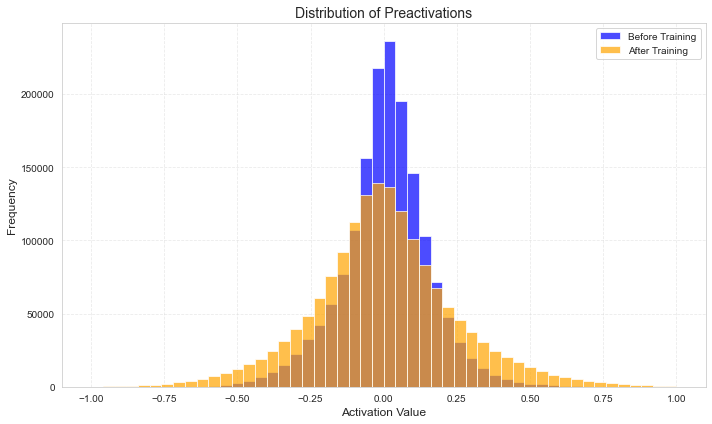}
    \caption{Histogram of preactivation values from a hidden layer in ResNet50, before and after training with \method{}. The post-training distribution is more concentrated and symmetric, reflecting improved activation stability.}
    \label{fig:distribution_preactivations}
\end{figure}

\autoref{fig:distribution_preactivations} illustrates the distribution of preactivations before and after training ResNet50 with \method{}. The post-training distribution is more concentrated and symmetric, indicating that \method{} stabilizes activation dynamics and mitigates internal covariate shift.

\subsection{Smoothness and Adaptability}

\begin{figure}[t]
    \centering
    \begin{subfigure}[b]{0.11\textwidth}
        \centering
        \includegraphics[width=\textwidth]{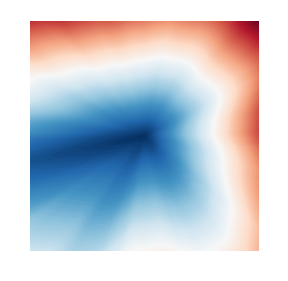}
        \caption{ReLU}
        \label{fig:relu_smoothness}
    \end{subfigure}
    \begin{subfigure}[b]{0.11\textwidth}
        \centering
        \includegraphics[width=\textwidth]{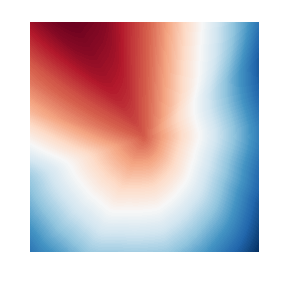}
        \caption{Swish}
        \label{fig:swish_smoothness}
    \end{subfigure}
    \begin{subfigure}[b]{0.11\textwidth}
        \centering
        \includegraphics[width=\textwidth]{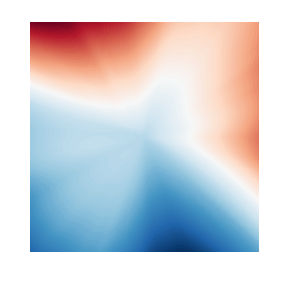}
        \caption{GELU}
        \label{fig:gelu_smoothness}
    \end{subfigure}
    \begin{subfigure}[b]{0.11\textwidth}
        \centering
        \includegraphics[width=\textwidth]{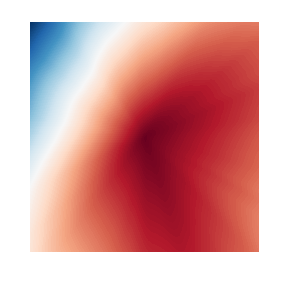}
        \caption{\method{}}
        \label{fig:velu_smoothness}
    \end{subfigure}
    \caption{Output landscapes of a randomly initialized ResNet50 with different activation functions. \method{} yields a smoother, variance-adaptive landscape compared to ReLU, Swish, and GELU.}
    \label{fig:smoothness_comparison}
\end{figure}

Because all components in \autoref{eq:velu} are smooth, \method{} is continuously differentiable. This contrasts with ReLU, which is continuous but non-differentiable at zero. The first derivative of \method{} can be written (in simplified form) as

\begin{equation}
\label{eq:first_derivative}
\begin{aligned}
f'(x) &= 
\lambda\, \sigma'(h(x))\, h'(x)\, s_{\text{adaptive}}
\\
&\quad +\;
\lambda\, x\, \sigma(h(x))\,
\gamma\, m\, \mu \,
\sech^{2}\!\left( m\,\mu\,\sigma_x \right).
\end{aligned}
\end{equation}

where $h(x) = \alpha(\tan^{-1}(\beta_1 x) + \sin^{-1}(\beta_2 x))$, and $\sigma'(h(x)) = \sigma(h(x))(1-\sigma(h(x)))$. The $\sech^{2}(\cdot)$ term arises from the derivative of the $\tanh(m \mu \sigma_x)$ in \autoref{eq:ArcTan-Sin_formulation}; in practice, $\sigma_x$ is treated as slowly varying within a batch, making this expression a good approximation of the gradient behavior. This derivative shows that \method{} maintains non-zero gradients across a broad activation range, avoiding the sharp gradient sparsity seen in ReLU and the rapid saturation of sigmoid/tanh. The resulting loss landscape is smoother and better conditioned, as visualized in \autoref{fig:smoothness_comparison}, which improves optimization stability in deep networks.

By dynamically adapting to the input distribution and incorporating Wasserstein-2 regularization, \method{} maintains well-behaved activation distributions throughout training, reduces internal covariate shifts, and more stable convergence, as demonstrated empirically in \autoref{sec:exp}.

\subsection{Parameter Selection and Stability Considerations}
\label{sec:parameter_stability}

The behavior of \method{} is governed by the parameters 
$\alpha$, $\beta_{1}$, $\beta_{2}$, $\gamma$, $\text{momentum}$, and $\lambda_{OT}$, each controlling a distinct component of the activation: nonlinearity strength, curvature of the ArcTan-ArcSin transformation, adaptive scaling sensitivity, and regularization intensity. While \method{} is stable under the recommended default configuration, improper parameter scaling may lead to gradient explosion or numerical instability. This section formalizes the conditions under which \method{} behaves reliably.

\paragraph{Nonlinearity parameters ($\alpha$, $\beta_1$, $\beta_2$).}
Larger values sharpen the composite transformation 
$h(x)=\alpha(\tan^{-1}(\beta_1 x) + \sin^{-1}(\beta_2 x))$, which increases sensitivity to input variations.  
However, excessively steep curvature amplifies the term 
$\sigma'(h(x))h'(x)$ in the gradient, potentially producing exploding updates.  
To mitigate this, we recommend  
$0.05 \le \alpha \le 0.2$ and $0.05 \le \beta_1,\beta_2 \le 0.3$ for stable training across CNNs and ViTs.

\paragraph{Adaptive scaling parameter ($\gamma$).}
The batch-adaptive factor  
\[
a(x) = 1 + \gamma \cdot \tanh(\text{momentum} \cdot \sigma_x)
\]
modulates the activation shape according to batch dispersion.  
To avoid excessively large scaling we apply a clipped form:
\begin{equation}
\label{eq:adaptive_factor}
\begin{aligned}
\text{adapt\_factor}
&= \operatorname{clip}\!\left( 1 + \gamma\,\tanh(\text{std\_dev}\cdot\text{momentum}),\right. \\
&\qquad\left. 0.5,\; 2.0 \right),
\end{aligned}
\end{equation}

Experiments show that values $\gamma \le 0.2$ guarantee smooth adaptation without destabilizing backpropagation.

\paragraph{Numerical stability.}
The computation
\[
\text{std\_dev}=\big(\mathbb{E}[x^2]+\epsilon\big)^{-1/2}
\]
is sensitive to small batch variances; a too-small $\epsilon$ produces extreme scaling.  
We recommend $\epsilon=10^{-3}$ rather than $10^{-6}$.

\paragraph{Wasserstein regularization ($\lambda_{OT}$).}
$\lambda_{OT}$ controls the strength of distributional alignment to a Gaussian target.  
Large values bias the activations toward overly constrained distributions and may hinder optimization.  
Empirically stable regimes~\cite{shakarami2025stress} lie within  
$\lambda_{OT}\in[0.001,0.05]$.  
If divergence or NaN gradients occur, reducing $\lambda_{OT}$ immediately stabilizes training.

\paragraph{Practical guidelines.}
To ensure stable optimization across all architectures, we recommend:
\begin{itemize}
\item Use gradient clipping (e.g., \texttt{clipnorm = 1.0}).
\item Use $\epsilon=10^{-3}$ in variance computations.
\item Clamp the adaptive scaling factor to $[0.5,2.0]$.
\item Reduce $\lambda_{OT}$ when facing divergence.
\end{itemize}

These recommendations establish a principled and reproducible parameter regime for \method{}.

\subsection{Extended Theoretical Properties of \method{}}
\label{sec:extended_theory}
This section formalizes the differentiability, adaptive behavior, and regularization properties of \method{}. 

\subsection{Definition}
For an input $x\in\mathbb{R}$, \method{} is defined as:
\[
\mathrm{VeLU}(x)
= \lambda \, x \, \sigma(h(x)) \, a(x)
- \lambda_{OT}\, W_2,
\]
where $\sigma$ is the sigmoid function, $h(x)$ is the composite transformation, $a(x)$ is the adaptive scaling term, and $W_2$ is the empirical Wasserstein-2 alignment penalty.

\begin{equation}
\label{eq:h_and_a}
\begin{aligned}
h(x) &= 
\alpha \!\left[
    \tan^{-1}(\beta_1 x)
    + \sin^{-1}(\beta_2 x)
\right],
\\[4pt]
a(x) &= 
1 + \gamma\, 
\tanh\!\left( m\,\sigma_x \right).
\end{aligned}
\end{equation}

\subsubsection{Gradient}
Using the product rule:
\begin{equation}
\label{eq:velu_grad_full}
\begin{aligned}
\frac{\partial\,\mathrm{VeLU}}{\partial x}
&=
\lambda\!\left[
    \sigma(h(x))
    + x\,\sigma'(h(x))\,h'(x)
\right] a(x)
\\[4pt]
&\quad
+\, \lambda\, x\, \sigma(h(x))\, a'(x)
\;-\;
\lambda_{OT}\, \frac{\partial W_2}{\partial x}.
\end{aligned}
\end{equation}

with:
\[
h'(x)=\alpha\!\left(
\frac{\beta_1}{1+(\beta_1 x)^2}
+
\frac{\beta_2}{\sqrt{1-(\beta_2 x)^2}}
\right).
\]
All components are smooth on their respective domains, ensuring $\mathrm{VeLU}\in\mathcal{C}^1(\mathbb{R})$.

\subsubsection{Adaptive Scaling Interpretation}
The scaling term satisfies:
\[
\lim_{\sigma_x\to 0} a(x) = 1,\qquad
\lim_{\sigma_x\to\infty} a(x) = 1+\gamma.
\]
Thus \method{} increases nonlinearity in high-variance regimes while reducing curvature in low-variance regimes, providing stable and context-aware activation shaping.

\subsubsection{Wasserstein Alignment}

The OT regularizer minimizes:
\[
W_2^2(\mathbb{P}_x,\mathcal{N}(0,1))
= \mathbb{E}\|x - T(x)\|^2,
\]
where $T$ is the optimal transport map.  
This encourages stable, Gaussian-like activation distributions, reducing internal covariate shift and improving generalization.

These properties together ensure that \method{} maintains smooth gradients, adapts dynamically to feature statistics, and regularizes activations through principled distributional alignment.

\subsection{Component-wise Functional Roles}
To understand the functional contribution of each design element in \method{}, we conceptually isolate the individual components and examine their theoretical role in shaping the activation behavior. The base multiplicative term $x$ preserves identity-like behavior near the origin and ensures that gradient flow is not hindered for small inputs. Adding the ArcTan-ArcSin composite introduces smooth, bounded curvature that enhances non-linearity while avoiding the abrupt saturation typical of hard-threshold functions.

Incorporating the variance-adaptive scaling term enables the activation to modulate its effective shape in response to the dispersion of preactivations, making \method{} sensitive to local activation statistics without introducing additional learnable parameters. Finally, the Wasserstein-2 regularization term aligns the activation's output distribution with a stable Gaussian target, reducing activation drift and promoting well-conditioned feature representations throughout training.

Together, these components form a progressively refined activation mechanism: the ArcTan-ArcSin structure shapes the non-linearity, the adaptive scaling provides data-dependent flexibility, and the OT regularization contributes distributional stability. The full \method{} formulation therefore integrates curvature, adaptivity, and statistical alignment into a unified activation framework.

\section{Experiments}
\label{sec:exp}

We evaluated \method{} by comparing it with standard activation functions, including ReLU~\cite{nair2010rectified}, Swish~\cite{ramachandran2017searching, tan2019efficientnet}, ReLU6~\cite{howard2017mobilenets}, and GELU~\cite{hendrycks2016gaussian,dosovitskiy2020image}, across multiple deep learning architectures and image benchmarks. Unless otherwise stated, we kept all training hyperparameters (optimizer, learning rate schedule, batch size, and data augmentation) identical across activations. Key experiments were repeated with multiple random initializations, and we report mean performance; variance statistics are provided in the supplementary material.

\label{scenarios}
We considered two integration scenarios: (1) \emph{partial replacement} of activation functions and (2) \emph{full replacement} of a model's activation functions with \method{}. In the first scenario, \method{} was used only in the classification head (the final layers), while the feature extraction backbone retained its original activations. This setup assesses compatibility with existing backbones and is particularly relevant for fine-tuning pre-trained models~\cite{yosinski2014transferable}. Most experiments, including the architectural and benchmark comparisons, follow this scenario. In contrast, the experiments in \autoref{tab:velu_resolution} and \autoref{tab:velu_optimizers}, as well as the results summarized in \autoref{tab:resnet50_velu_accuracy} and \autoref{fig:resnet50_velu_comparison}, implement the second scenario, where all activation functions in ResNet50~\cite{he2016deep} (backbone and added layers) are replaced by \method{}. This demonstrates that \method{} can serve as the primary activation function in a network without relying on additional nonlinearities.

For convolutional models (e.g., ResNet~\cite{he2016deep}, VGG~\cite{simonyan2014very}, MobileNet~\cite{howard2017mobilenets}, and DenseNet~\cite{huang2017densely}), we used a global average pooling layer~\cite{lin2013network} before the classifier. For ViT\_B16~\cite{dosovitskiy2020image}, which outputs a 2D feature vector, we omitted pooling. The classifier head consisted of two fully connected layers with 256 and 64 neurons, each followed by an activation function, and a final softmax layer. This standardized design enables a fair comparison of activations across architectures with different inductive biases.

We conducted experiments on CIFAR-10 and CIFAR-100~\cite{krizhevsky2009learning}, MNIST~\cite{lecun1998gradient}, Fashion-MNIST~\cite{xiao2017fashion}, Corel-1K, and Corel-10K~\cite{wang2001simultaneous}. Although these benchmarks are relatively classical, they remain standard for evaluating activation functions due to their controlled complexity, ease of reproducibility, and broad adoption in related work. Notably, recent activation proposals such as Swish~\cite{ramachandran2017searching} and GELU~\cite{hendrycks2016gaussian} also rely on these datasets, enabling meaningful comparisons.

For CIFAR, MNIST, and Fashion-MNIST, we followed the original training/validation splits of the benchmarks. For Corel-1K and Corel-10K, we randomly divided the data into 80\% training and 20\% test sets. Each model was trained for 5-20 epochs, depending on network depth and convergence, and we report mean top-1/top-5 accuracy, loss, training time, inference throughput (FPS), and memory usage.

Although we provide default initialization values for \method{} in our public implementation\footnote{\url{https://github.com/AshkanShakarami/Velu}}, these can be adjusted for specific tasks. This flexibility enables task-dependent optimization~\cite{bengio2012practical}, improved generalization~\cite{goodfellow2016deep}, and adaptation to different architectures~\cite{tan2019efficientnet}. However, manual tuning can be time-consuming~\cite{bergstra2012random, feurer2019hyperparameter} and computationally demanding~\cite{li2017hyperband}. As a direction for future work, we propose integrating automated hyperparameter optimization (e.g., Bayesian optimization~\cite{snoek2012practical}, gradient-based meta-learning~\cite{finn2017model}, or search frameworks such as Optuna~\cite{akiba2019optuna}, Hyperopt~\cite{bergstra2013hyperopt}, and KerasTuner~\cite{omalley2019kerastuner}) to set suitable initial values based on sampled data, thereby improving usability and reducing the need for manual tuning.

\subsection{Evaluation Across Architectures}

\autoref{tab:architecture_comparison} compares \method{} with standard activation functions across a range of architectures, including CNNs~\cite{lecun1998gradient, krizhevsky2012imagenet, shakarami2023tcnn} and ViTs~\cite{dosovitskiy2020image}. Across all models in this setting (partial replacement in the classifier head), \method{} achieves higher top-1 accuracy than the corresponding baseline activation, confirming its adaptability to both convolutional and transformer-based designs.

Performance gains are particularly pronounced in deeper or residual networks such as ResNet50 and DenseNet121, where \method{} improves top-1 accuracy by nearly 2\%, suggesting better gradient flow and more stable optimization. In lightweight architectures like MobileNetV2, the gains are smaller but still positive, indicating suitability under parameter- and compute-constrained conditions. ViT-B16 also benefits: \method{} slightly outperforms GELU in both top-1 and top-5 accuracy, showing compatibility with self-attention mechanisms and LayerNorm-based pipelines. These trends indicate that \method{} delivers generalizable improvements across architectures with different depth, connectivity, and inductive biases, supporting its role as a plug-in alternative to standard activations.

\begin{table}[t]
\centering
\caption{Architectural performance with and without \method{} (partial replacement). Top-1 and Top-5 accuracy (\%).}
\label{tab:architecture_comparison}

\setlength{\tabcolsep}{4pt}

\begin{tabular}{lccc}
\toprule
\textbf{Model} & \textbf{Act.} & \textbf{Top-1} & \textbf{Top-5} \\
\midrule

\multirow{2}{*}{EffNetB3}  
    & \method{} & \textbf{87.61} & \textbf{99.40} \\
    & Swish & 87.18 & 99.22 \\
\midrule

\multirow{2}{*}{VGG19}  
    & \method{} & \textbf{90.97} & 99.71 \\
    & ReLU & 90.30 & \textbf{99.82} \\
\midrule

\multirow{2}{*}{ResNet50}  
    & \method{} & \textbf{91.27} & 99.41 \\
    & ReLU & 89.44 & \textbf{99.61} \\
\midrule

\multirow{2}{*}{MobileNetV2}  
    & \method{} & \textbf{91.32} & 99.70 \\
    & ReLU6 & 90.69 & \textbf{99.77} \\
\midrule

\multirow{2}{*}{DenseNet121}  
    & \method{} & \textbf{91.85} & \textbf{99.49} \\
    & ReLU & 89.99 & 99.38 \\
\midrule

\multirow{2}{*}{ViT-B16}  
    & \method{} & \textbf{96.25} & \textbf{99.95} \\
    & GELU & 95.80 & 99.87 \\
\bottomrule
\end{tabular}

\end{table}

\subsection{Evaluation Across Benchmarks}

\autoref{tab:baseline_results} compares MobileNetV2~\cite{sandler2018mobilenetv2} with and without \method{} across 12 datasets. Integrating \method{} consistently improves top-1 accuracy, with the largest gains observed on CIFAR-100 and the Corel datasets, which involve more classes and higher intra-class variability.

In CIFAR-100 and Corel-10K, \method{} improves both top-1 and top-5 accuracy, indicating better feature discrimination in more complex classification settings. On simpler benchmarks such as MNIST and Fashion-MNIST, where baseline performance is already near-saturated, the improvements are smaller but still positive in top-1 accuracy. Overall, these results show that \method{} enhances performance across tasks of varying complexity.

\begin{table}[t]
\centering
\caption{MobileNetV2 performance with and without \method{} (partial replacement). Top-1 and Top-5 accuracy (\%).}
\label{tab:baseline_results}

\begin{tabular}{lccc}
\toprule
\textbf{Benchmark} & \textbf{Model} & \textbf{Top-1} & \textbf{Top-5} \\
\midrule

\multirow{2}{*}{CIFAR-10}  
    & Baseline & 90.69 & \textbf{99.77} \\
    & \method{} & \textbf{91.32} & 99.70 \\
\midrule

\multirow{2}{*}{CIFAR-100}  
    & Baseline & 71.05 & 91.54 \\
    & \method{} & \textbf{72.63} & \textbf{92.36} \\
\midrule

\multirow{2}{*}{MNIST}  
    & Baseline & 99.01 & 99.99 \\
    & \method{} & \textbf{99.21} & \textbf{100} \\
\midrule

\multirow{2}{*}{Fashion-MNIST}  
    & Baseline & 91.10 & \textbf{99.75} \\
    & \method{} & \textbf{91.13} & 99.50 \\
\midrule

\multirow{2}{*}{Corel1K}  
    & Baseline & 91.00 & 99.00 \\
    & \method{} & \textbf{93.50} & \textbf{100} \\
\midrule

\multirow{2}{*}{Corel10K}  
    & Baseline & 86.45 & 95.70 \\
    & \method{} & \textbf{87.90} & \textbf{97.50} \\
\bottomrule
\end{tabular}
\end{table}

\autoref{tab:resnet50_velu_accuracy} reports top-1 accuracy for ResNet50~\cite{he2016deep} with full activation replacement by \method{} (Scenario~2), along with the absolute gain over the original activation. Here, \method{} yields consistent improvements across all benchmarks. The largest gains occur on Corel-1K (+9.00\%), Corel-10K (+3.85\%), and CIFAR-100 (+3.12\%), reflecting stronger benefits on more challenging, multi-class datasets. On simpler datasets such as MNIST and Fashion-MNIST, the absolute gains are smaller but still positive.

\begin{table}[t]
\centering
\caption{ResNet50 performance with full \method{} activation replacement (Scenario~2). Accuracy and absolute gains.}
\label{tab:resnet50_velu_accuracy}
\begin{tabular}{lcc}
\toprule
\textbf{Benchmark} & \textbf{Acc (\%)} & \textbf{Gain} \\
\midrule
CIFAR-10        & 91.29 & \textbf{+1.90} \\
CIFAR-100       & 72.07 & \textbf{+3.12} \\
MNIST           & 99.34 & \textbf{+0.36} \\
Fashion-MNIST   & 92.26 & \textbf{+0.66} \\
Corel1K         & 93.50 & \textbf{+9.00} \\
Corel10K        & 87.35 & \textbf{+3.85} \\
\bottomrule
\end{tabular}
\end{table}

\begin{figure*}[t]
    \centering
    \includegraphics[width=0.95\textwidth]{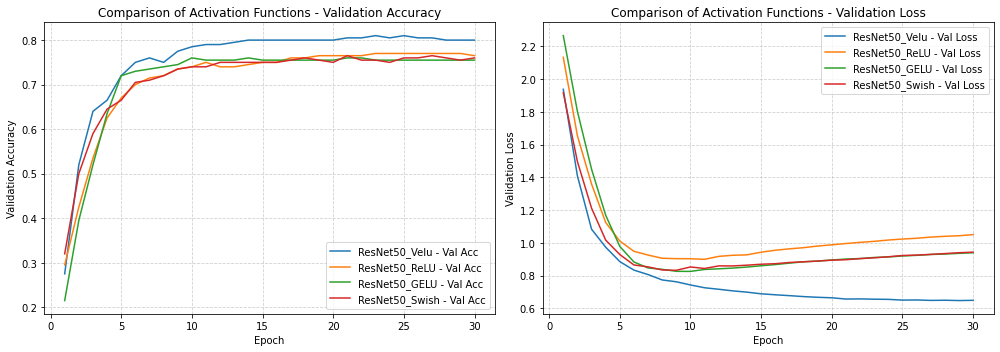}
    \caption{%
    Comparison of \method{} with ReLU, Swish, and GELU in ResNet50.
    (\textit{Left}) Validation accuracy: \method{} surpasses all baselines after only a few epochs and keeps a consistent advantage throughout training.
    (\textit{Right}) Validation loss: \method{} shows a steep early drop and converges to the lowest loss, whereas the baselines plateau earlier and ReLU exhibits a mild late increase.
    }
    \label{fig:resnet50_velu_comparison}
\end{figure*}

\autoref{fig:resnet50_velu_comparison} presents the validation accuracy and loss curves for ResNet50 trained with \method{} compared to ReLU, GELU, and Swish. Already after the first few epochs (around epoch~2--3), \method{} reaches higher validation accuracy than all baselines and maintains this advantage throughout training. This indicates that \method{} not only converges quickly with few epochs, but also preserves a consistent edge in the mid- and late-training phases.

The validation-loss plot shows a similar trend: \method{} exhibits one of the steepest early loss drops and then continues to decrease smoothly, stabilizing at the lowest loss level. In contrast, the losses for ReLU, GELU, and Swish flatten earlier, with ReLU even showing a mild late-epoch increase suggestive of overfitting. These dynamics are consistent with the variance-aware design of \method{}, which supports both fast convergence - beneficial for transfer learning and fine-tuning scenarios - and more stable, well-conditioned optimization in later training stages.

\subsection{Training Efficiency and Memory Usage}

\autoref{tab:architecture_performance} compares activations in terms of final training loss, training time, inference speed, and memory usage (partial replacement scenario). In all convolutional architectures, \method{} achieves lower training loss than the corresponding baseline activation, indicating more effective optimization. In ViT-B16, GELU retains a slightly lower loss, but \method{} remains competitive.

In terms of training time and FPS, \method{} is comparable to the baselines. It is slightly faster in some architectures (e.g., MobileNetV2 and ResNet50) and slightly slower in others (e.g., EfficientNetB3, DenseNet121, ViT-B16), reflecting backend- and architecture-dependent kernel optimizations rather than fundamental complexity gaps. Memory usage with \method{} is also close to that of the baselines, with modest increases in some models due to the additional functional composition.

Overall, \autoref{tab:architecture_performance} shows that \method{} delivers consistent gains in training loss and accuracy while maintaining competitive training time, inference throughput, and memory footprint, making it practical for deployment even in resource-constrained settings.

\begin{table}[t]
\centering
\caption{Performance of various architectures with and without \method{}. Loss (L), Training Time (T), FPS, and Memory based on Megabyte (M) Usage (MB).}
\label{tab:architecture_performance}

\begin{tabular}{lccccc}
\toprule
\textbf{Model} & \textbf{Act.} & \textbf{L} & \textbf{T (s)} & \textbf{FPS} & \textbf{M} \\
\midrule

\multirow{2}{*}{EffNetB3}  
    & \method{} & \textbf{0.42} & 2621.03 & \textbf{16.07} & 14786 \\
    & Swish & 0.44 & \textbf{2417.54} & 15.16 & \textbf{13403} \\
\midrule

\multirow{2}{*}{VGG19}  
    & \method{} & \textbf{0.30} & \textbf{2047.57} & 25.50 & \textbf{14581} \\
    & ReLU & 0.37 & 2113.78 & \textbf{28.65} & 14730 \\
\midrule

\multirow{2}{*}{ResNet50}  
    & \method{} & \textbf{0.39} & 1362.65 & \textbf{20.20} & 13705 \\
    & ReLU & 0.53 & \textbf{1353.11} & 19.76 & \textbf{13119} \\
\midrule

\multirow{2}{*}{MobileNetV2}  
    & \method{} & \textbf{0.33} & \textbf{819.40} & \textbf{21.15} & 14440 \\
    & ReLU6 & 0.37 & 844.61 & 16.08 & \textbf{13871} \\
\midrule

\multirow{2}{*}{DenseNet121}  
    & \method{} & \textbf{0.32} & 2246.35 & 15.78 & \textbf{13628} \\
    & ReLU & 0.43 & \textbf{1751.09} & \textbf{17.62} & 13777 \\
\midrule

\multirow{2}{*}{ViT-B16}  
    & \method{} & 0.14 & 692.38 & 11.29 & \textbf{13259} \\
    & GELU & \textbf{0.13} & \textbf{676.85} & \textbf{12.33} & 13554 \\
\bottomrule
\end{tabular}
\end{table}

\subsection{Behavior on Different Image Resolutions}

\autoref{tab:velu_resolution} reports MobileNetV2 performance with full \method{} replacement on CIFAR-10 at different input resolutions. \method{} maintains stable and improving performance as resolution increases: from 75.10\% top-1 accuracy at $32 \times 32$ up to 93.34\% at $224 \times 224$, with top-5 accuracy consistently above 98.5\%. This robustness suggests that \method{} is suitable for settings where input resolution varies, such as transfer learning~\cite{tan2019efficientnet}, medical imaging~\cite{litjens2017survey, shen2017deep}, and real-time applications~\cite{howard2019searching}.

\begin{table}[t]
\centering
\caption{Accuracy of \method{} on CIFAR-10 across different input resolutions using MobileNetV2 (full activation replacement).}
\label{tab:velu_resolution}
\begin{tabular}{lcc}
\toprule
\textbf{Resolution} & \textbf{Top-1} & \textbf{Top-5} \\
\midrule
32$\times$32     & 75.10  & 98.51 \\
64$\times$64     & 86.51  & 99.45 \\
128$\times$128   & 91.32  & 99.70 \\
224$\times$224   & 93.34  & 99.82 \\
\bottomrule
\end{tabular}
\end{table}

\subsection{Adaptability With Different Optimizers}

\autoref{tab:velu_optimizers} evaluates \method{} (full replacement) on ResNet50 with different optimizers on CIFAR-10 at $128 \times 128$ resolution. Adam achieves the highest top-1 accuracy (91.32\%), but SGD, Nadam, and RMSProp all maintain high top-5 accuracy, demonstrating that \method{} can be trained effectively under different optimization dynamics. This robustness to optimizer choice is valuable in practice, where the preferred optimizer may depend on computational constraints, convergence behavior, or prior workflow choices~\cite{kingma2014adam, ruder2016overview, hinton2012rmsprop}.

\begin{table}[t]
\centering
\caption{Accuracy of \method{} as a full replacement for ResNet50 activations on CIFAR-10 across different optimizers at $128\times128$ resolution.}
\label{tab:velu_optimizers}

\begin{tabular}{lcc}
\toprule
\textbf{Optimizer} & \textbf{Top-1} & \textbf{Top-5} \\
\midrule
Adam      & 91.32 & 99.70 \\
SGD       & 89.33 & 99.70 \\
Nadam     & 87.23 & 99.19 \\
RMSProp   & 86.95 & 99.03 \\
\bottomrule
\end{tabular}
\end{table}

\section{Extended Experiments}
\label{sec:extended_results}

\subsection{Robustness Across Image Resolutions}
\label{sec:resolution_analysis}

To evaluate resolution-invariance, we trained MobileNetV2 with \method{} on CIFAR-10 at four input sizes.  
As shown in ~\autoref{tab:velu_resolution}, \method{} maintains stable performance across all resolutions, achieving consistent gains as image size increases.  
Notably, even at $32\times32$ resolution - where many adaptive activations underperform - \method{} retains competitive accuracy.  
This supports our claim that the variance-adaptive scaling mitigates resolution-induced activation drift.

\subsection{Optimizer Sensitivity Analysis}

We next examined how \method{} interacts with different optimization algorithms.  
Results for ResNet50 on CIFAR-10 (~\autoref{tab:velu_optimizers}) show that \method{} improves performance under all four optimizers tested, with the largest gains under Adam.  
SGD and RMSProp also maintain high Top-5 accuracy, confirming that \method{} does not rely on a specific optimizer to achieve stable convergence.

\subsection{Evaluation on Additional Benchmarks}
\begin{table}[t]
\centering
\caption{Performance comparison of the Baseline (B.) model with and without \method{} in different vision benchmarks. The original Baseline results are referenced from \cite{sandler2018mobilenetv2}.}
\label{tab:appendix_results}
\begin{tabular}{lccc}
\toprule
\textbf{Benchmark} & \textbf{Model} & \textbf{Top-1} & \textbf{Top-5} \\
\midrule
\multirow{2}{*}{PathMNIST~\cite{yang2023medmnist}}  
    & B. & 95.01 & 99.90 \\
    & \textbf{\method{} in B} & \textbf{95.31} & 99.90 \\
\midrule
\multirow{2}{*}{BloodMNIST~\cite{yang2023medmnist}}  
    & B. & 96.61 & 99.94 \\
    & \textbf{\method{} in B.} & \textbf{97.08} & \textbf{100} \\
\midrule
\multirow{2}{*}{PACS~\cite{zhang2023towards}}  
    & B. & 97.70 & 100 \\
    & \textbf{\method{} in B.} & \textbf{98.10} & 100 \\
\midrule
\multirow{2}{*}{EuroSAT-RGB~\cite{helber2019eurosat}}  
    & B. & 96.11 & \textbf{99.96} \\
    & \textbf{\method{} in B.} & \textbf{96.43} & 99.93 \\
\midrule
\multirow{2}{*}{Oxford-IIIT Pet~\cite{cubuk2018autoaugment}}  
    & B. & 81.92 & 96.61 \\
    & \textbf{\method{} in B.} & \textbf{83.62} & \textbf{97.77} \\
\midrule
\multirow{2}{*}{Tiny-ImageNet~\cite{li2021boosting}}  
    & B & 69.31 & 87.12 \\
    & \textbf{\method{} in B.} & \textbf{70.43} & \textbf{87.74} \\
\bottomrule
\end{tabular}
\end{table}

To assess generalization across domains and difficulty levels, we evaluated \method{} on medical, satellite, domain-generalization, and fine-grained classification datasets. Shown in ~\autoref{tab:appendix_results},  
consistent improvements across PathMNIST, BloodMNIST, PACS, EuroSAT, Oxford-IIIT Pet, and Tiny-ImageNet demonstrate that \method{} enhances learning in both low- and high-variance regimes. Importantly, improvements on medical datasets confirm that the Wasserstein-based alignment stabilizes representations in small-sample settings, while gains on Tiny-ImageNet and Oxford-IIIT Pet indicate improved fine-grained discrimination. Similar neural descriptor stability is critical in CAD systems across neuroimaging and pathology~\cite{shakarami2020cad,shakarami2021diagnosing,shakarami2025unit}.

\section{Limitations and Future Work}

\paragraph{Limitations.} Although the experimental results demonstrate that the proposed activation design achieves consistent improvements across various architectures and benchmarks, several limitations remain. First, the current study focuses primarily on medium-scale vision datasets, and large-scale evaluations on more diverse domains are still required to fully assess the activation's generalizability. Second, while we conducted extensive testing across CNNs and vision transformers, the behavior of the activation in sequence models, multimodal frameworks, and reinforcement-learning settings remains unexplored. Third, our implementation relies on a fixed configuration strategy; a systematic analysis of automatic parameter tuning, stability ranges, and hyperparameter interactions would further clarify the operational boundaries of the method.

Another limitation lies in the scope of ablations: although we investigated the contribution of major components, additional controlled experiments are needed to disentangle interactions between adaptive modulation, normalization aspects, and distributional regularization. Moreover, while computational overhead remains moderate, future work should examine kernel-level optimizations and hardware-specific acceleration for deployment in resource-constrained environments.

\paragraph{Future Work.}
Alongside extending evaluations to larger datasets and more diverse architectures, we are actively developing an enhanced version of the activation function which incorporates additional mechanisms aimed at improving stability, adaptability, and distributional control without introducing architectural burden. Details of this extension will be presented in a dedicated publication. Beyond this, several research directions appear promising: (i) integrating the activation into foundation models and emerging transformer variants; (ii) exploring its compatibility with low-precision and quantized training; (iii) analyzing theoretical convergence properties under different optimization regimes; and (iv) investigating its role as a building block in self-normalizing or specific training pipelines.

Together, these directions highlight the potential for continued refinement of activation functions that adapt to feature statistics and maintain stable representational dynamics at scale.

\section{Conclusion}
We introduced \method{}, a variance-aware activation function that combines ArcTan–ArcSin transformations, adaptive scaling, and Wasserstein-2 regularization. Across CNNs and ViTs, \method{} consistently improved accuracy and stability over ReLU, Swish, and GELU, while requiring only a single learnable parameter and minimal computational overhead. Although our experiments focused on medium-scale vision benchmarks, the results indicate that variance-adaptive and distribution-aligned activations are a promising direction. Future work will include large-scale evaluations and exploring \method{} in Transformer-based and sequence models.

\end{document}